%% file: main.tex
\documentclass[runningheads]{llncs}
\usepackage{graphicx}

\usepackage{hyperref}

\usepackage{cite}  %
\usepackage{marvosym}  %
\input{math_commands.tex}

\usepackage{url}

\newcommand{\dd}[2]{\frac{\partial #1} {\partial #2}}
\newcommand{\der}[2]{\frac{d #1} {d #2}}

\newcommand{\Ham}{\mathcal{H}}

\begin{document}
\title{Learning Trajectories of Hamiltonian Systems with Neural Networks}
\author{Katsiaryna Haitsiukevich\textsuperscript{\Letter} \and Alexander Ilin}
\authorrunning{K. Haitsiukevich, A. Ilin}

\institute{Aalto University, Espoo, Finland\\
\email{\{firstname.lastname\}@aalto.fi}}
\maketitle              %
\begin{abstract}

Modeling of conservative systems with neural networks is an area of active research. A popular approach is to use Hamiltonian neural networks (HNNs) which rely on the assumptions that a conservative system is described with Hamilton's equations of motion. Many recent works focus on improving the integration schemes used when training HNNs. In this work, we propose to enhance HNNs with an estimation of a continuous-time trajectory of the modeled system using an additional neural network, called a deep hidden physics model in the literature. We demonstrate that the proposed integration scheme works well for HNNs, especially with low sampling rates, noisy and irregular observations.

\end{abstract}
\section{Introduction}
Many real-world physical systems are modeled using (partial) differential equations which are derived from the laws of physics. This modeling approach has the benefits that one can build a functional model using a small amount of data and the model may generalize well outside of the training data distribution (provided that the modeling assumptions are correct). However, building the model from the first principles requires deep understanding of the modeled process and often results in a tedious procedure when various modeling assumptions are tested in how well they can explain the data.
The data-driven approach is therefore an attractive alternative: one can fit a generic model like neural networks to training data without much effort on the model design and the derivations of the learning and inference procedures. The downside is, however, that the accuracy of this model depends greatly on the amount of available data: too little data may result in models that do not generalize well.
Thus, there is clear demand for combining two modeling approaches: using neural networks models for better flexibility while also constraining the solutions with laws of physics can greatly improve sample efficiency.
The laws of conservation (of energy/mass/momentum) are among very common modeling assumptions made for describing physical systems. Many real-world physical systems can be modeled as closed and therefore conservative systems.
Combining the conservation laws with neural networks (see, e.g., \cite{greydanus2019hamiltonian,cranmer2020lagrangian,hoedt2021mclstm,jagtap2020conservative,Lee2021deepconservation}) is therefore a promising line of research with many potential applications.

One prominent research direction that emerged recently in the literature is modeling Hamiltonian systems with neural networks \cite{greydanus2019hamiltonian}. In Hamiltonian neural networks (HNNs), the law of the energy conservation is in-built in the structure of the dynamics model and therefore it is automatically satisfied. The idea of utilizing Hamilton's equations was successfully used to predict the dynamics of Hamiltonian systems from pixel observations \cite{toth2020hamiltonian,hochlehnert2021physicallystruct,zhong2020unsupervised}, to build representations of molecular data \cite{li2021hamnet} and it was extended to control tasks \cite{zhong2020symoden,zhong2020unsupervised} and meta-learning~\cite{lee2021identifying}.
The original HNN model \cite{greydanus2019hamiltonian} had the limitation of assuming the knowledge of the state derivatives with respect to time or approximating those using finite differences.
Many recent works have used numerical integrators for modeling the evolution of the system state and several improvements of the integration procedure have been proposed \cite{chen2020srnn,dipietro2020ssinn,jin2020sympnets,tong2021taylornets,xiong2021nssnn,david2021shnn}.

In this paper, we propose to model the evolution of the system state by adding another neural network instead of relying on traditional numerical integrators. This alternative to numerical integration is known in the literature under the name \emph{deep hidden physics models} \cite{raissi2018dhp}. Our method provides a continuous-time approximation of system states without relying on additional assumptions such as Hamiltonian separability. Hamiltonian preservation is encoded as a soft constraint through an extra loss term rather than being in-built in the architecture itself. We experimentally show that the proposed approach can improve the modeling accuracy in the presence of observation noise and for measurements with low sampling rates or irregularly-sampled observations.

\section{Modeling Hamiltonian systems}

\subsection{Hamiltonian neural networks}

The modeling assumption of Hamiltonian neural networks
\cite{greydanus2019hamiltonian} is that the observed state $\vs=(\vq, \vp)$ of a dynamical system evolves according to Hamilton's equations:
\begin{align}
    \frac{d \mathbf{q}}{d t} = \frac{\partial\mathcal{H}}{\partial \mathbf{p}}
    \,,
    \qquad
    \frac{d \mathbf{p}}{d t} = -\frac{\partial\mathcal{H}}{\partial \mathbf{q}}
    \,,
\label{eq:hamilt}
\end{align}
where $\Ham$ is the Hamiltonian (total energy) of the system, $\vq$ is the position and $\vp$ is the momentum part of the state. Hamiltonian equations assume the system state $\vs=(\vq, \vp)$ being represented in canonical coordinates. Although Hamiltonian $\Ham(\vq, \vp)$ is modeled with a generic neural network with inputs $\vq$ and $\vp$, using \eqref{eq:hamilt} to describe the system dynamics guarantees that the total energy is conserved:
\begin{align}
  \dd{\Ham}{t} = \dd{\Ham}{\vp} \der{\vp}{t} + \dd{\Ham}{\vq} \der{\vq}{t} = 0 .
\end{align}

The original HNN model \cite{greydanus2019hamiltonian} was trained by minimizing the loss 
\begin{align}
\mathcal{L}_{\text{HNN}} = \frac{1}{K} \sum_{k=1}^{K} \left(\frac{d\vq_k}{dt} - \dd{\mathcal{H}_k}{\vp}\right)^2
+ \left(\frac{d\vp_k}{dt} + \dd{\mathcal{H}_k}{\vq}\right)^2
\label{eq:hnnloss}
\end{align}
where $\dd{\Ham_k}{\vp}$, $\der{\vp_k}{t}$, $\dd{\Ham_k}{\vq}$, $\der{\vq_k}{t}$ are partial derivatives computed at the locations of the training examples $\vq(t_k), \vp(t_k)$. The derivatives $\dd{\Ham_k}{\vp}$, $\dd{\Ham_k}{\vq}$ are calculated by differentiating the neural network which models the Hamiltonian, while the derivatives $\der{\vp_k}{t}$, $\der{\vq_k}{t}$ are either assumed to be known (from the simulator) or approximated with finite differences.
Using finite differences to approximate the derivatives $\der{\vp}{t}$ and $\der{\vq}{t}$ is essentially equivalent to Euler integration with a time step being equal to the sampling interval, which limits the accuracy of the trained model \cite{david2021shnn}.

Many extensions of HNNs \cite{toth2020hamiltonian,chen2020srnn,dipietro2020ssinn,tong2021taylornets} use more advanced numerical integrators combined with the Neural ODE approach \cite{chen2018neuralode} to model the evolution of the system state in time. Several works \cite{chen2020srnn,dipietro2020ssinn,tong2021taylornets,jin2020sympnets,xiong2021nssnn} use symplectic integrators which preserve the conserved quantity and therefore are natural options for Hamiltonian systems. The analysis \cite{zhu2020deephnn} of several numerical integrators when applied to HNNs shows that non-symplectic integrators cannot guarantee the recovery of true Hamiltonian~$\mathcal{H}$ and the prediction accuracy obtained with a symplectic integrator depends on the integrator accuracy order. In addition to the use of a symplectic integrator, SympNets \cite{jin2020sympnets} %
have the network architecture that guarantees zero energy loss by network design. 
Some symplectic integration schemes \cite{chen2020srnn,dipietro2020ssinn,tong2021taylornets} make additional assumptions such as the separability of the Hamiltonian.\footnote{Hamiltonian $\Ham(\vq, \vp) = V(\vq) + K(\vp)$ where $V$ and $K$ are potential and kinetic energies is separable.} A recent model called Non-separable Symplectic Neural Networks (NSSNN) \cite{xiong2021nssnn} releases this assumption by an improved symplectic integrator which works well for both separable and non-separable Hamiltonians.

One potential problem in applying model equations~\ref{eq:hamilt} to real-world data is the fact that the system dynamics in \eqref{eq:hamilt} is written for clean states $\vp$ and $\vq$ while in practice state measurements typically contain noise. Working with noisy states leads to inaccurate modeling due to the compounding error problem. Therefore the states should be denoised both at training and inference times. Many existing HNN models were trained with noisy observations but they do not have in-built techniques for handling observation noise in initial states at inference time. Some works directly address this issue: for example, \cite{chen2020srnn} which proposes an initial state optimization procedure.

\subsection{Physics-informed neural networks as an integrator for HNN}
\label{sec:pinn}

Physics-informed neural networks (PINNs) \cite{lagaris1998artificial,raissi2019pinn} is a mesh-free method of solving given differential equations using neural networks.
The method can be used to approximate a solution of an initial value problem defined by an ordinary differential equation (ODE) with a \emph{known} function $\vf$
\begin{equation}
    \frac{d \vs (t)}{d t} = \vf(t, \vs(t))
\label{eq:pde}
\end{equation}
and initial conditions $\vs_{0} = \vs(t_0)$.
The solution $\vs(t)$ of the ODE is approximated by a neural network with time $t$ as input trained by minimizing a composite loss function. The loss function for the network training can be represented as a weighted sum of the supervised learning loss for the initial conditions
\begin{align}
  \mathcal{L}_\text{init} = ||\vs(t_0) - \vs_{0}||^2
\label{eq:ibcond}
\end{align}
and the loss forcing the network to satisfy the ODE in \eqref{eq:pde}:
\begin{align}
  \mathcal{L}_\text{ode} =
  \frac{1}{K} \sum_{k=1}^{K}
  \left(\frac{d \vs_k}{d t} - \vf(t_k)\right)^2
\label{eq:lossode}
\end{align}
where $\der{\vs_k}{t}$ denotes derivatives computed at locations $t_k$. The locations $t_k$ can be sampled randomly on the interval on which the ODE is solved.

In case a sequence of observations $\{(t_0, \vs_0), ..., (t_N, \vs_N)\}$ from the modelled system in~\eqref{eq:pde} is available PINNs allow to easily include the observations in the model training procedure in which case the loss in \eqref{eq:ibcond} is replaced with the following supervision loss:
\begin{align}
  \mathcal{L}_\text{fit} = \frac{1}{N} \sum_{i=1}^N ||\vs(t_i) - \vs_{i}||^2.
\label{eq:losssup}
\end{align}
The PINN method can be viewed as a supervised learning method with the ODE-based regularizer given in \eqref{eq:lossode}. Unlike traditional numerical solvers PINNs can handle ill-posed problems, e.g. with unknown initial conditions but with measurements for other time points.

Deep hidden physics models (DHPMs) \cite{raissi2018dhp} extend the PINNs approach to the case of an \emph{unknown} function $\vf$ in \eqref{eq:pde}.
Function $\vf$ is approximated with another neural network which is trained by minimizing the loss in \eqref{eq:lossode}. Thus, DHPMs contain two neural networks: one defines the differential equation and the other one approximates its solution.

In this paper, we propose to learn the HNN model using the PINNs approach, that is to approximate the solution of the Hamiltonian equations~\ref{eq:hamilt} by a neural network $\vs(t)$ that outputs $\vq(t)$ and $\vp(t)$ as a function of time $t$. This network is trained to fit the available observations $\vs_i=(\vq_i, \vp_i)$ at time instances $t_i$ by minimizing the loss in \eqref{eq:losssup} and to satisfy Hamilton's equations by minimizing the loss in \eqref{eq:hnnloss}. The value of Hamiltonian $\Ham$ is approximated by an HNN which is jointly trained with solution network $\vs(t)$.
Note that the derivatives $\der{\vq_k}{t}$, $\der{\vp_k}{t}$ can be computed by differentiating the neural network $\vs(t)$ with respect to its input $t$. Thus, the method does not require the knowledge of these derivatives from the simulator or their approximation using finite differences. 
Note also that the locations of points $t_k$ do not have to coincide with the locations of the training samples: they are sampled randomly on the solution interval.
This stabilizes training and makes the method work well with lower sampling rates and in the presence of observation noise.

The total loss minimized during training is the weighted sum of losses in \eqref{eq:hnnloss} and~\eqref{eq:losssup}.
We also find it beneficial to use an additional loss term which forces the energy values to stay constant throughout a trajectory:
\begin{equation}
    \mathcal{L}_\text{extra} =
\frac{1}{M} \sum_{ij} \left(\Ham(\vs(t_{i})) - \Ham(\vs(t_j)) \right)^2 \,,
\label{eq:lossextra}
\end{equation}
where pairs of points $t_i, t_j$ are sampled randomly and $M$ is the number of sampled pairs in a mini-batch. 

We call our algorithm Deep Hidden Hamiltonian (DHH) in analogy to DHPMs: we assume that Hamiltonian $\Ham$ is unknown and should be learned from data.

\section{Related work}

Recently many improvements to the original HNN architecture have been proposed in the literature.
The proposed improvements include the use of symplectic integrators \cite{chen2020srnn,dipietro2020ssinn,tong2021taylornets,jin2020sympnets,xiong2021nssnn}, hard constraints on energy conservation \cite{jin2020sympnets} as well as modifications to the soft constraints, for example, by switching to the Cartesian coordinates \cite{finzi2020simplifying}. More details on the comparison of different methods for Hamiltonian systems can be found in survey~\cite{zhong2021survey}. In contrast to many exiting works, our method does not rely on traditional numerical integrators but instead utilizes an extra neural network for learning of the system trajectory. Another alternative to approaches listed above is a method called GFNN \cite{chen2021gfnn} that learns modified generating functions as a symplectic map representation instead of approximating a vector field directly.
Thus, the method does not require finite difference approximations of the vector field.
However, GFNN requires solving a system of non-linear equations for prediction of one step evolution which might be a computational bottleneck.

An advantage of the algorithm proposed in this paper is its applicability to noisy measurements without the need to design a separate denoising procedure.
The solution network finds a continuous-time approximation of the system state trajectory from noisy observations. Previous works had a separate denoising step, for example, by basis functions approximations \cite{wu2020structure} or an optimization procedure for the initial step \cite{chen2020srnn}.

Our approach is related to several works that have used physics-informed neural networks for modeling conservative systems.
cPINNs \cite{jagtap2020conservative} has used physics-informed neural networks for modeling non-Hamiltonian conservative systems with known system equations. 
Work \cite{mattheakis2020hamiltonian} has applied PINNs for solving Hamilton's differential equations with a \emph{known} Hamiltionian $\Ham$. In contrast, we learn the Hamiltonian from the data following the DHPMs methodology.

\section{Experiments}

We test our method on the following four physical systems from \cite{greydanus2019hamiltonian}:
\begin{itemize}
    \item mass-spring
\begin{equation}
    \Ham = \frac{1}{2}k q^2 + \frac{p^2}{2m}
    \label{eq:spring_ham}
\end{equation}
    with $m=\frac{1}{2}$ and $k=2$;
    \item pendulum
\begin{equation}
    \Ham = 2mgl (1 - \cos q) + \frac{l^2p^2}{2m}
\end{equation}
    with $l=1$, $m=\frac{1}{2}$ and $g=3$;
    \item 2-body and 3-body systems
\begin{equation}
    \Ham = \sum_{i=0}^N \frac{\| \mathbf{p}_i\|^2}{2 m_i} - \sum_{1 \le i < j \le N} \frac{G m_i m_j}{\| \mathbf{q}_i - \mathbf{q}_j \|}
    \label{eq:nbody_ham}
\end{equation}
with $G=1$, $m_i=1$, $i = 1, \dots, N$ and $N \in \{2, 3\}$.
\end{itemize}
We generate training data by randomly sampling the initial state and numerically solving differential equations~\ref{eq:hamilt} using the fourth-order Runge-Kutta method with $\Ham$ given in equations~\ref{eq:spring_ham} --~\ref{eq:nbody_ham}. To test the robustness of the proposed method to noise in observations, the generated data points are corrupted with additive Gaussian noise with zero mean and $0.1$ standard deviation.

In all the experiments, Hamiltonian $\mathcal{H}$, solutions $\vs(t)$ and dynamics $\vf$ (for a DHPM baseline) are modeled with multi-layer perceptron networks. The models are optimized using the Adam optimizer \cite{kingma2017adam} with learning rates of $0.0001$ for HNN and dynamics $\vf$ and $0.01$ for the solution network.
We normalize the modeling interval of $t$ to be $[-1, 1]$. Points $t_k$ required for computing the loss terms in \eqref{eq:hnnloss} and \eqref{eq:lossode} for the DHPM baseline (described in Sec.~\ref{sec:pinn}) are sampled uniformly from the interval $[-1, 1]$ for each optimization step. Similarly, pairs of points $(t_i, t_j)$ for the extra loss term in \eqref{eq:lossextra} are sampled at each optimization step such that $t_i \in [-1, 0]$ and $t_j \in [0, 1]$. In our experiments, the training loss is the weighted sum of the supervision loss in \eqref{eq:losssup} (with weight 1), the loss in \eqref{eq:hnnloss} (with weight $0.1$ for the mass-spring system and the pendulum and $1$ for the 2-body and 3-body systems) and the loss in \eqref{eq:lossextra} (with weight $0.01$).

\begin{figure}[t]
\centering
\begin{minipage}[b]{.48\linewidth}
\centering
\includegraphics
[width=60mm,height=30mm,trim={7mm 3mm 10mm 12mm},clip]
{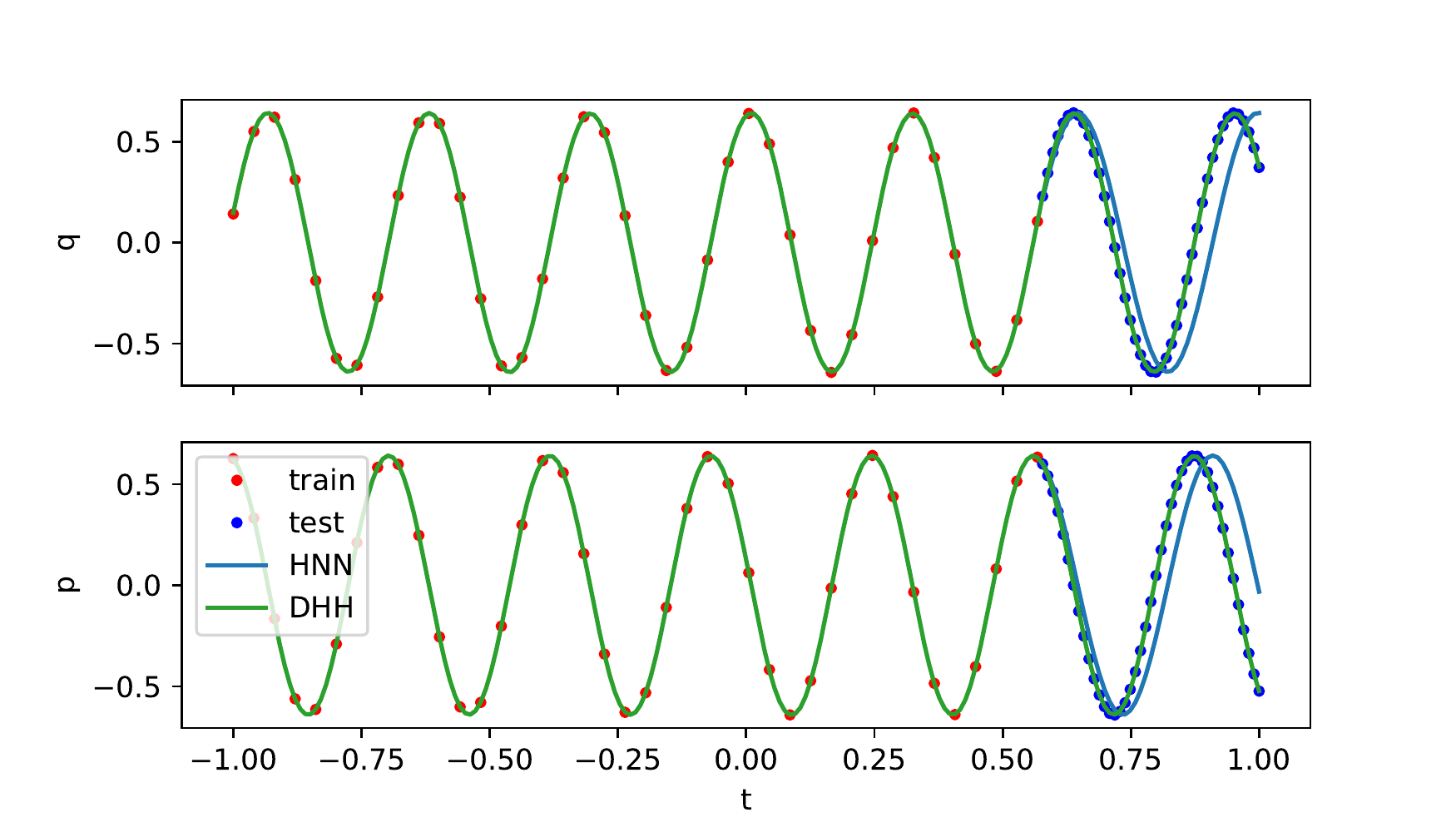}
\\
(a) clean observations
\end{minipage}
\begin{minipage}[b]{.48\linewidth}
\centering
\includegraphics
[width=60mm,height=30mm,trim={12mm 3mm 10mm 12mm},clip]
{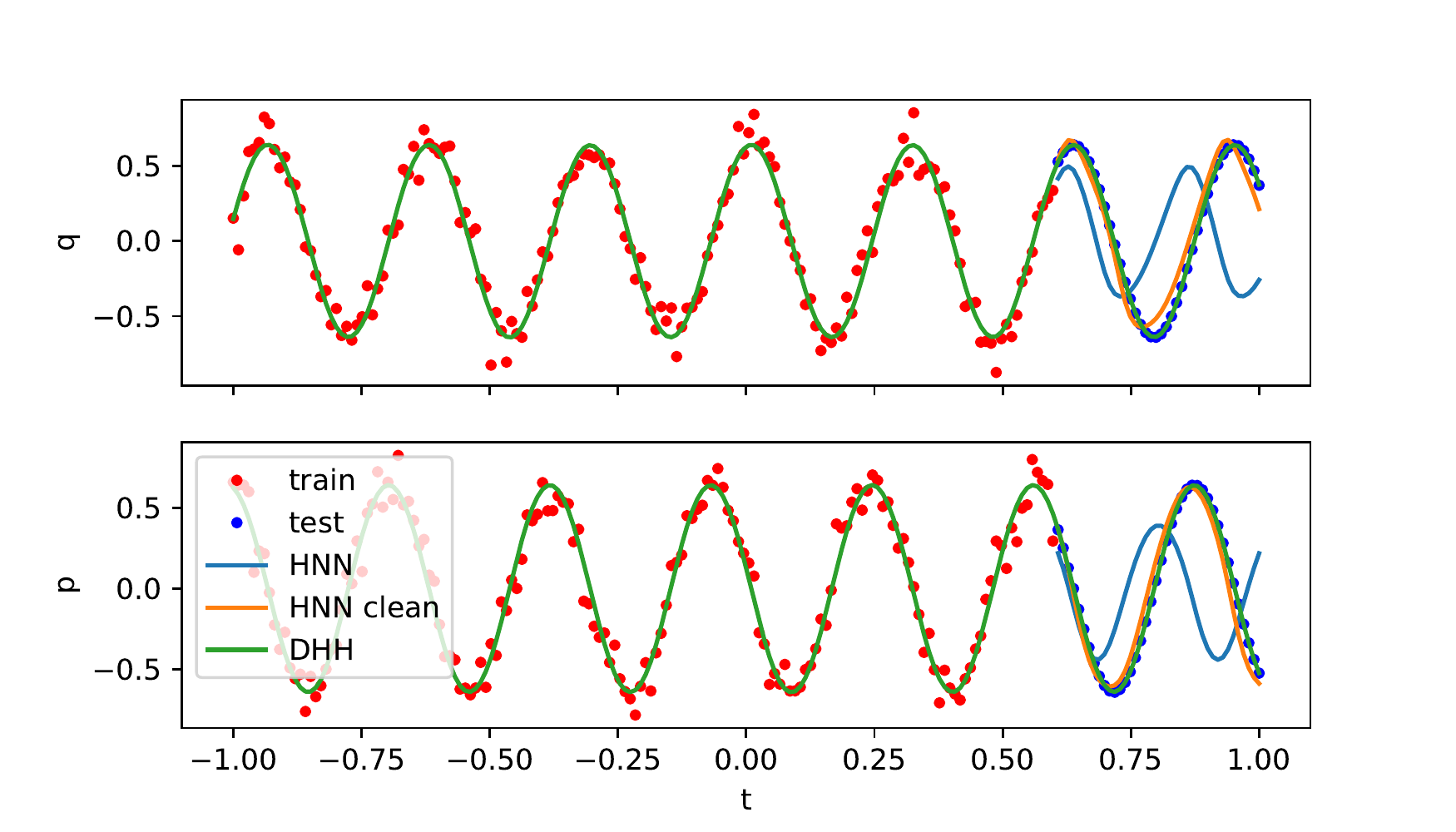}
\\
(b) noisy observations
\end{minipage}
\\[3mm]
\begin{minipage}[b]{.48\linewidth}
\centering
\includegraphics
[width=60mm,height=30mm,trim={7mm 3mm 10mm 12mm},clip]
{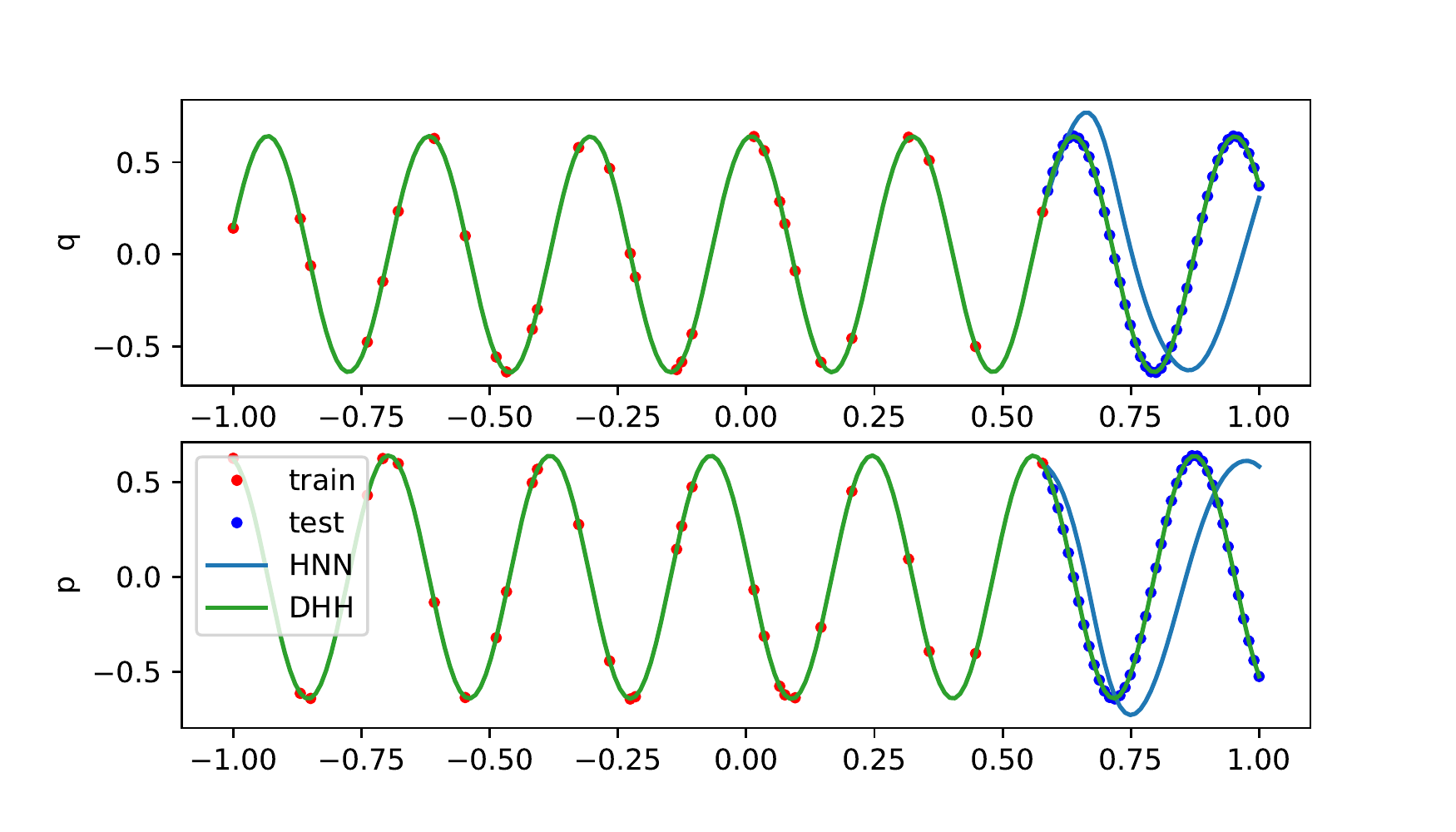}
\\
(c) clean irregularly-sampled observations
\end{minipage}
\begin{minipage}[b]{.48\linewidth}
\centering
\includegraphics
[width=60mm,height=30mm,trim={12mm 3mm 10mm 12mm},clip]
{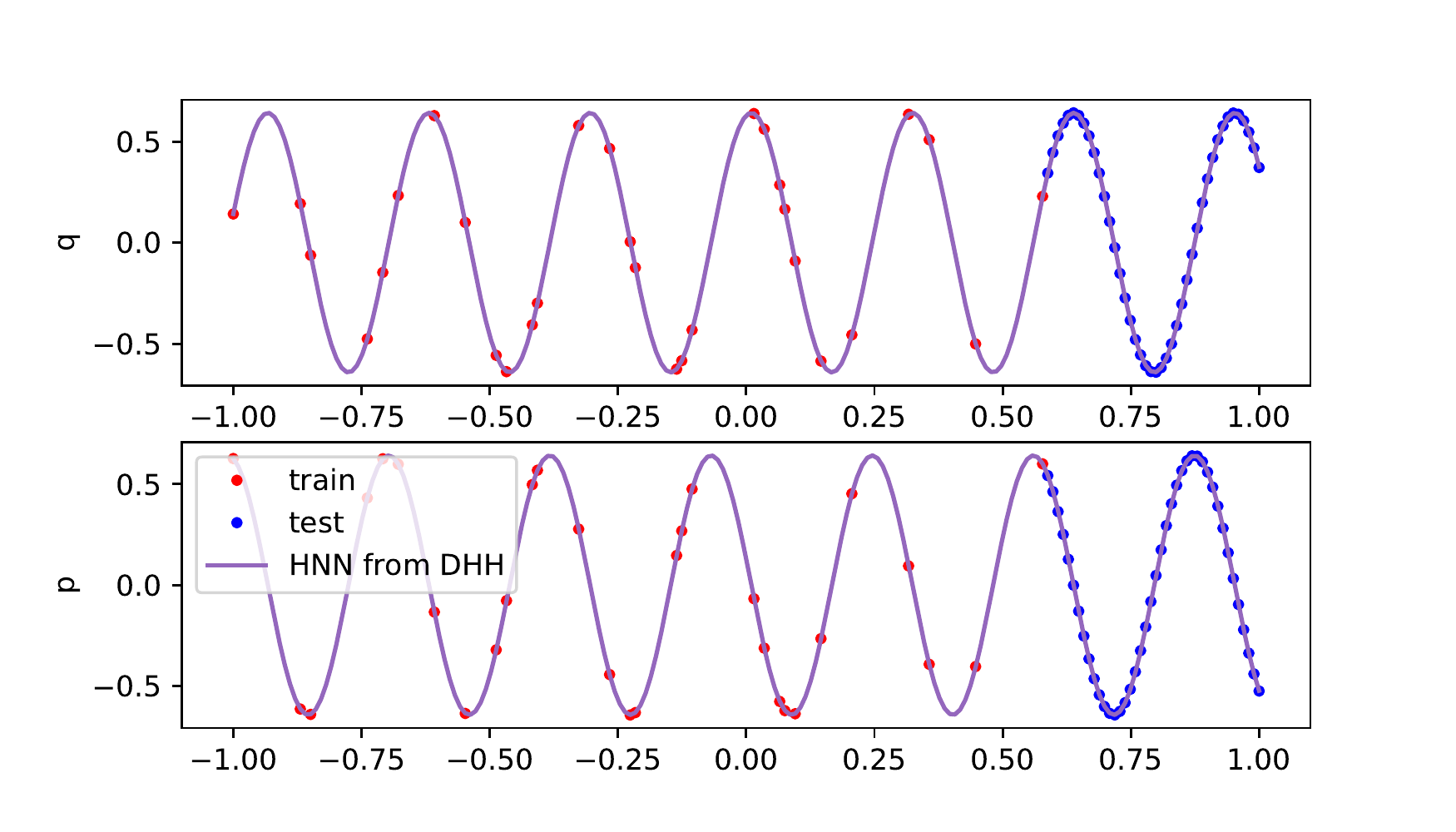}
\\
(d) DHH + Runge-Kutta integrator \\ at test time
\end{minipage}

\caption{(a)--(c): Estimated state trajectories  for the mass-spring system with DHH (green) and HNN (blue). (d): The trajectory obtained by integrating the system equations~\ref{eq:hamilt} with the Runge-Kutta integrator using the Hamiltonian $\Ham$ learned by DHH for the data from Fig.~\ref{fig:spring_traj}c.}
\label{fig:spring_traj}
\end{figure}

In Fig.~\ref{fig:spring_traj}, we compare the trajectories for the mass-spring system estimated with the proposed DHH approach and with the HNN from \cite{greydanus2019hamiltonian} in which the derivatives are estimated using finite differences. Fig.~\ref{fig:spring_traj}a-c show the results for three settings: 1)~clean regularly-sampled observations 2)~noisy regularly-sampled observations 3)~clean irregularly-sampled observations.
In the experiment with noisy observations (Fig.~\ref{fig:spring_traj}b), we additionally show a trajectory obtained by the baseline HNN when we assume access to the clean state observation for the last time step in the training set.
The results show that the proposed method performs well under low sampling rates and it tolerates noise in the data.

In Fig.~\ref{fig:spring_traj}d, we show the trajectory obtained by integrating the system equations~\ref{eq:hamilt} with the Euler integrator using the Hamiltonian $\Ham$ learned by DHH for the data from Fig.~\ref{fig:spring_traj}c. The results show that the Hamiltonian $\Ham$ found by DHH is very accurate and the model works well even when changing the integration scheme at test time.

\begin{figure}
\centering
\begin{minipage}[b]{.5\linewidth}
\includegraphics
[width=60mm,height=30mm,trim={7mm 3mm 10mm 12mm},clip]
{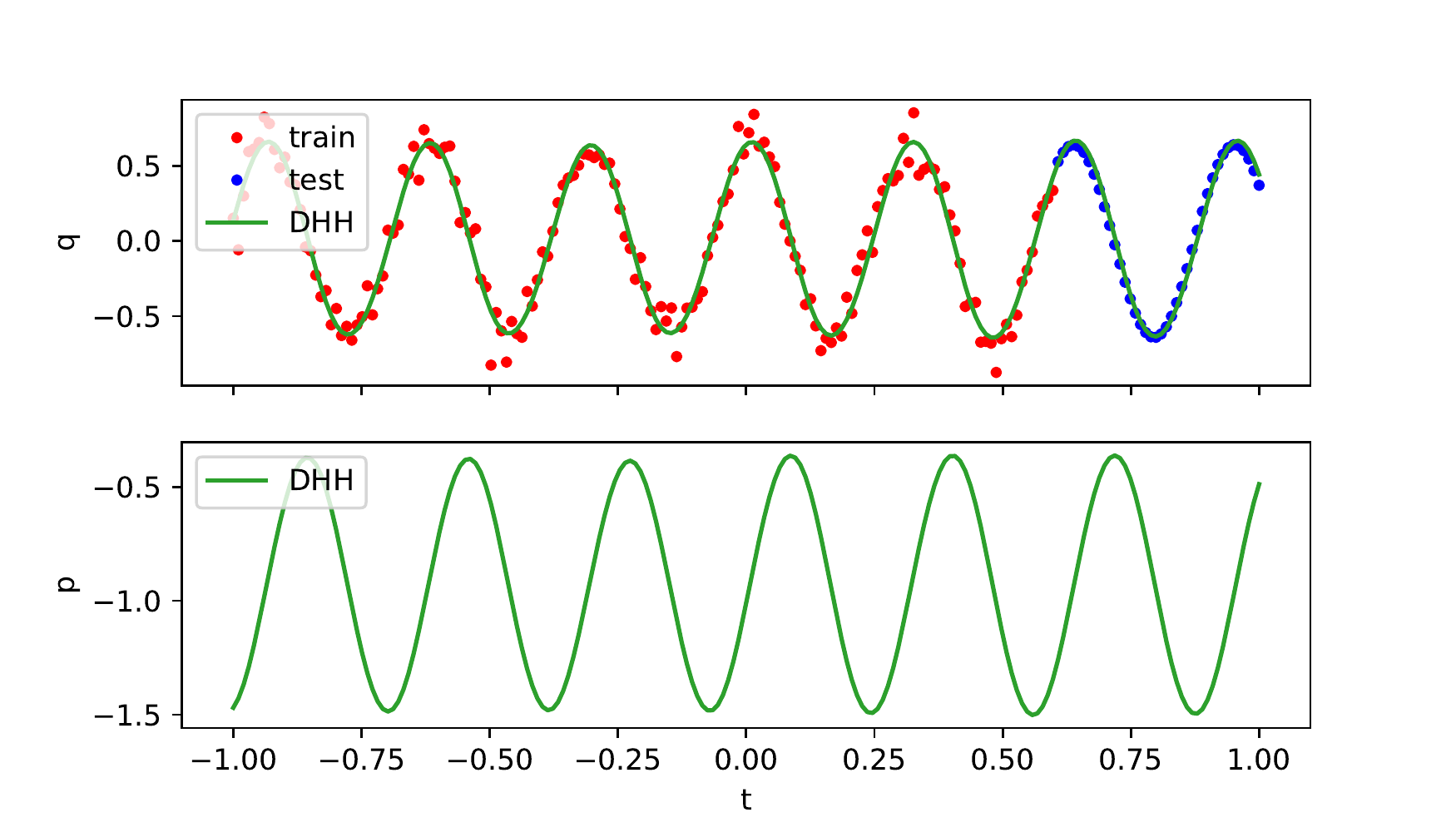}
\end{minipage}
\begin{minipage}[b]{.3\linewidth}
\includegraphics
[height=30mm,trim={0 0mm 3 10mm},clip]
{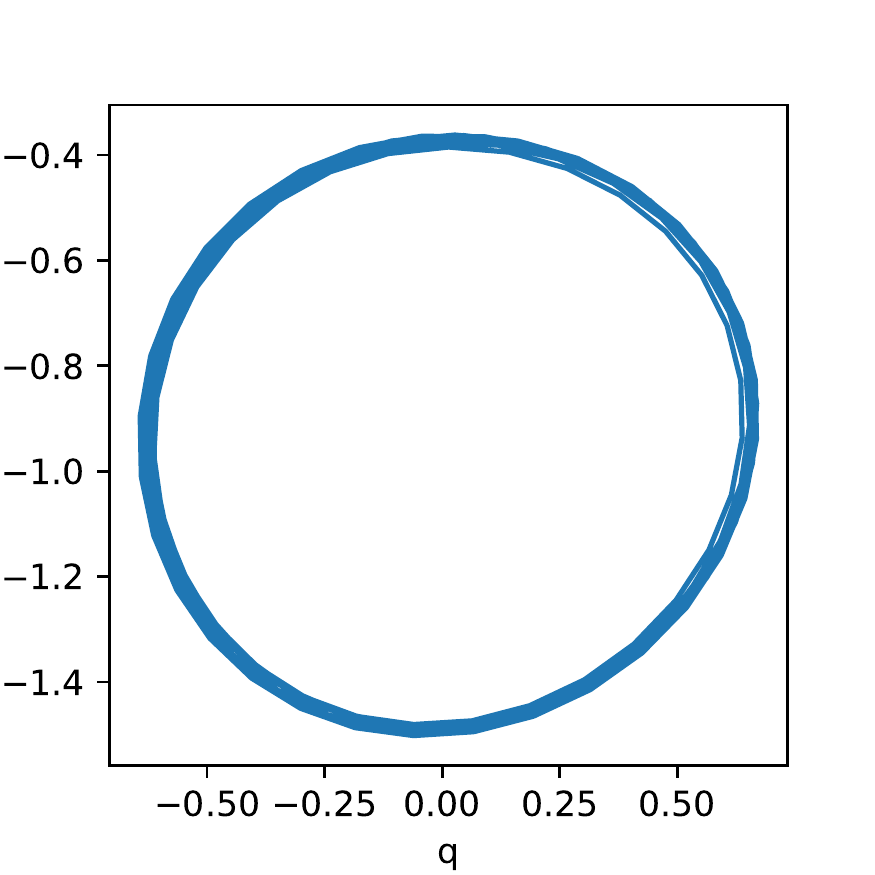}
\end{minipage}

\caption{Estimated trajectories and the corresponding vector field by DHH for the mass-spring system with noisy observations of $\vq$ and no observations of $\vp$.}
\label{fig:spring_partial}
\end{figure}

In Fig.~\ref{fig:spring_partial}, we show that the proposed algorithm can be used when some of the state variables are not observed. In this experiment, we model the mass-spring system using only noisy measurements of the position variable $\vq$ while $\vp$ is assumed unobserved. %
The results show that the proposed method is able to reconstruct the missing coordinate (up to an additive constant).

Next, we quantatively compare the proposed approach against the following baselines: 1)~HNN~\cite{greydanus2019hamiltonian} with derivatives calculated as finite differences, 2)~HNN~\cite{greydanus2019hamiltonian} with derivatives provided by the simulator, 3)~NSSNN \cite{xiong2021nssnn}, 4)~Neural ODE \cite{chen2018neuralode} and 5)~DHPMs \cite{raissi2018dhp}. The implementations of NSSNN and HNN are taken from the original papers. For Neural ODE, we use the same implementation as in \cite{xiong2021nssnn} with the second-order Runge-Kutta integrator.
DHPMs estimate the system dynamics in \eqref{eq:pde} by modeling $\vf$ with a multi-layer perceptron and by minimizing the sum of the losses in equations~\ref{eq:lossode} and~\ref{eq:losssup}.

\begin{figure}[ht]
\centering
\begin{minipage}{.24\linewidth}
\hspace{5mm}
\centering mass-spring
\end{minipage}
\hfill
\begin{minipage}{.24\linewidth}
\hspace{3mm}
\centering pendulum
\end{minipage}
\hfill
\begin{minipage}{.24\linewidth}
\hspace{3mm}
\centering 2-body system
\end{minipage}
\hfill
\begin{minipage}{.24\linewidth}
\hspace{2mm}
\centering 3-body system
\end{minipage}
\\
clean observations
\\[1mm]
\includegraphics[width=0.24\linewidth,trim={1mm 11mm 10mm 12mm},clip]{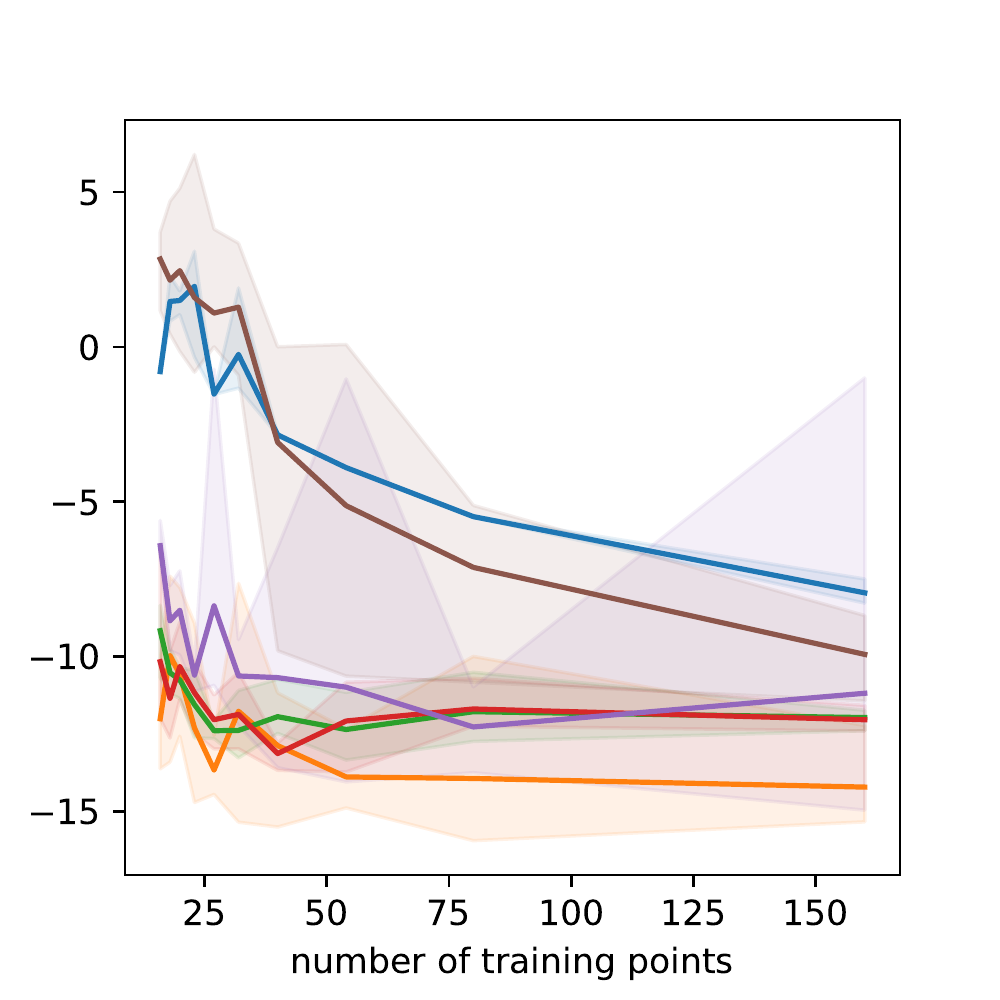}
\includegraphics[width=0.24\linewidth,trim={1mm 11mm 10mm 12mm},clip]{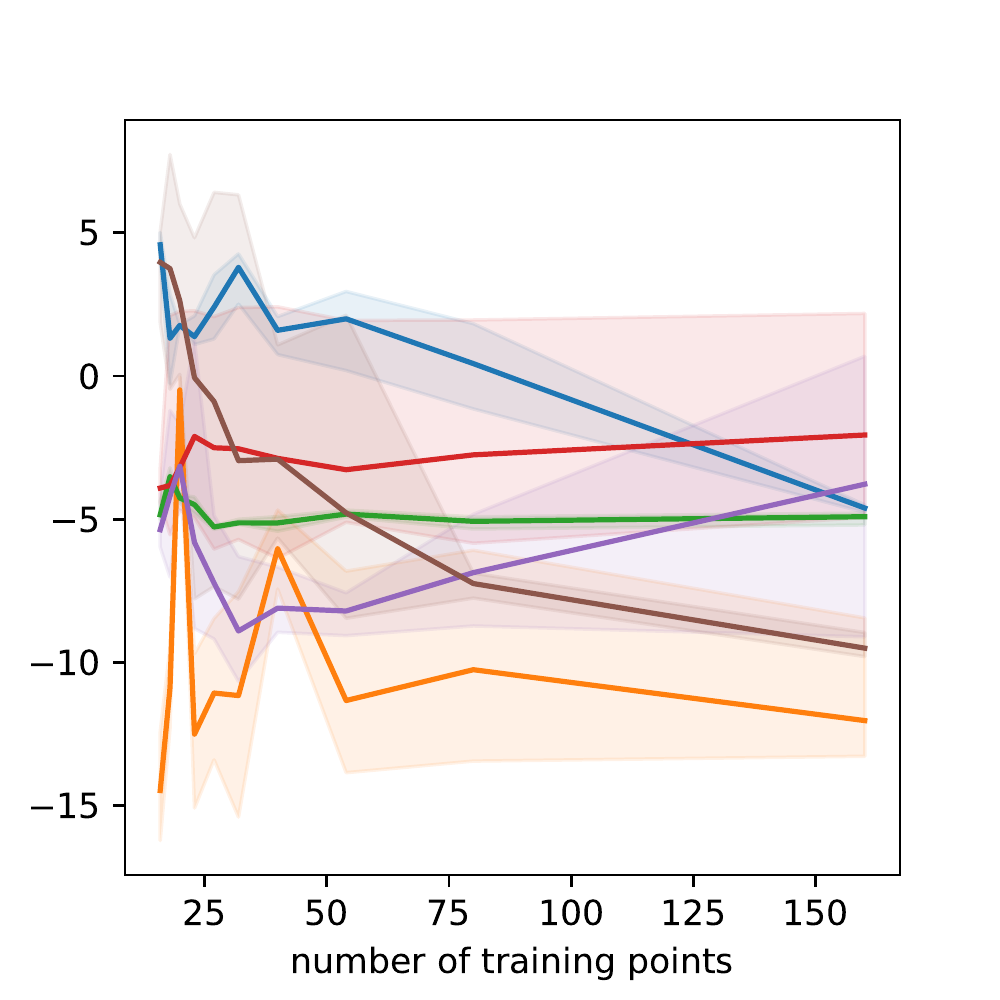}
\includegraphics[width=0.24\linewidth,trim={1mm 11mm 10mm 12mm},clip]{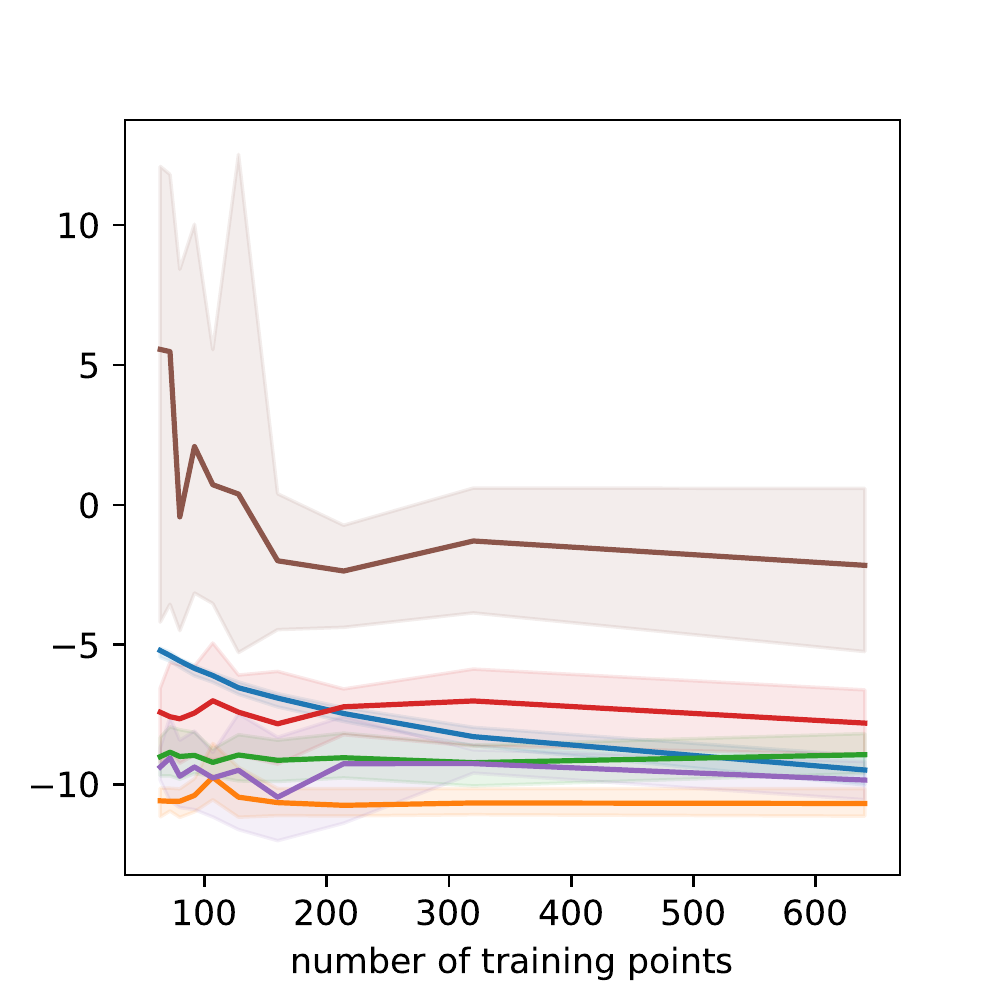}
\includegraphics[width=0.24\linewidth,trim={1mm 11mm 10mm 12mmm},clip]{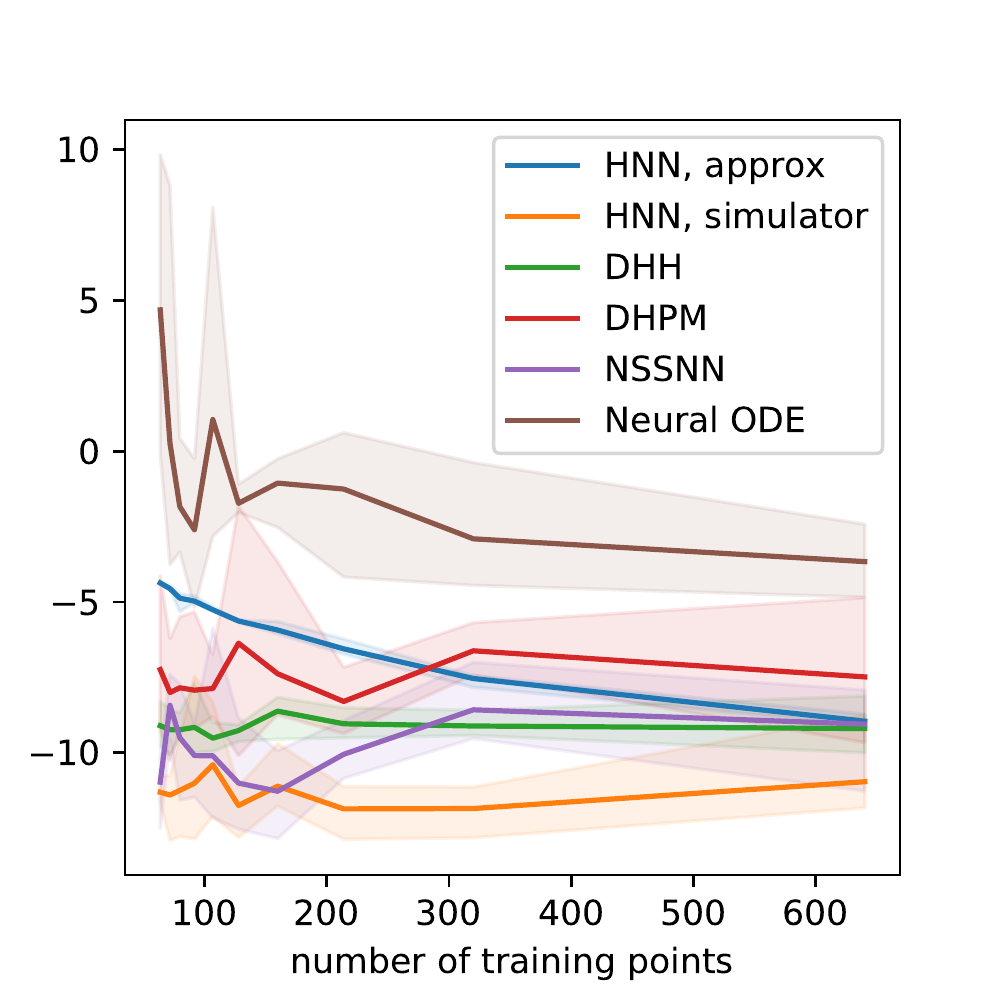}
\\
noisy observations
\\
\includegraphics[width=0.24\linewidth,trim={1mm 2mm 10mm 12mm},clip]{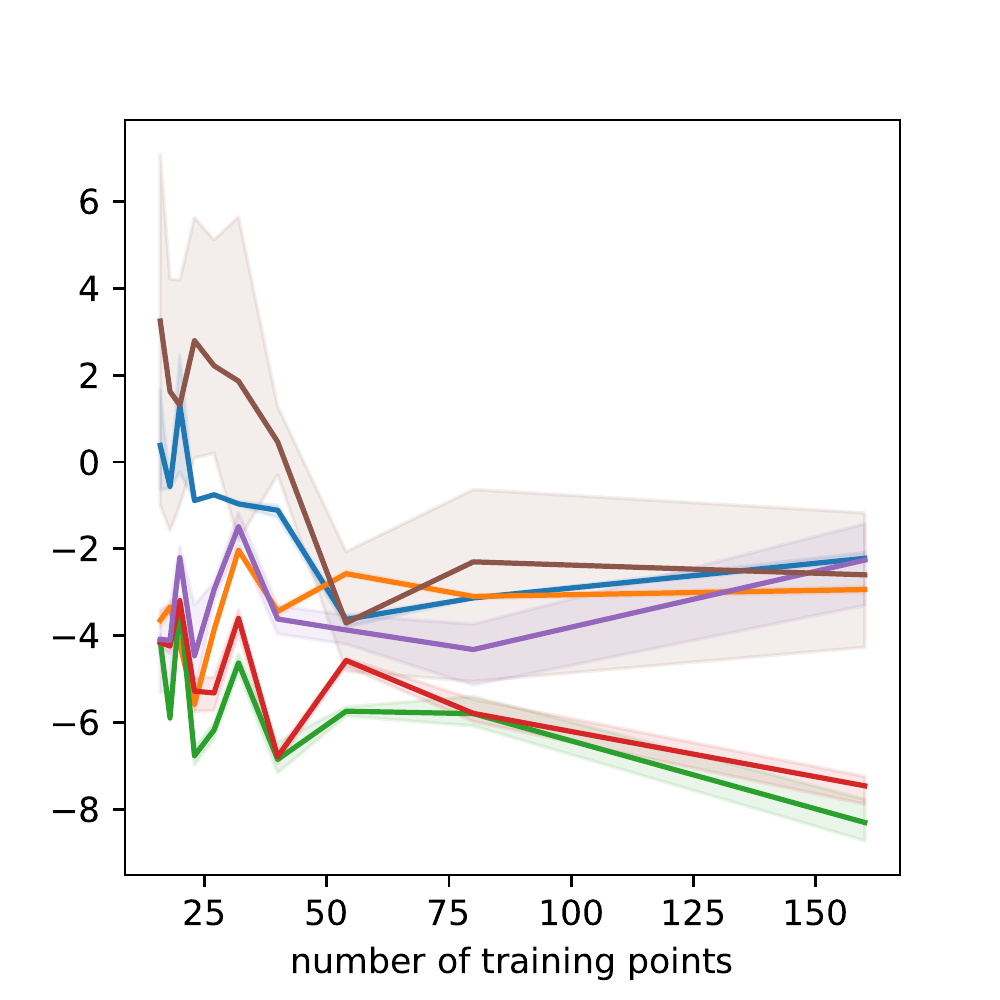}
\includegraphics[width=0.24\linewidth,trim={1mm 2mm 10mm 12mm},clip]{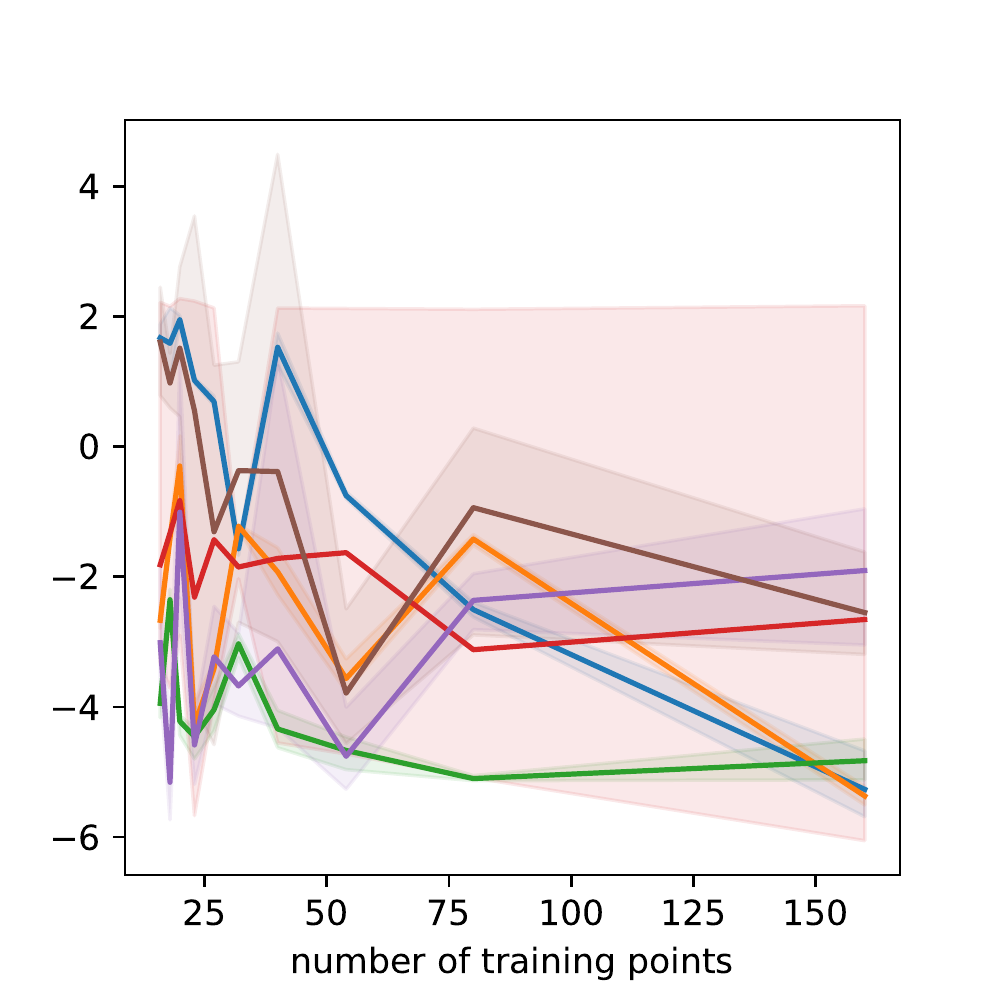}
\includegraphics[width=0.24\linewidth,trim={1mm 2mm 10mm 12mm},clip]{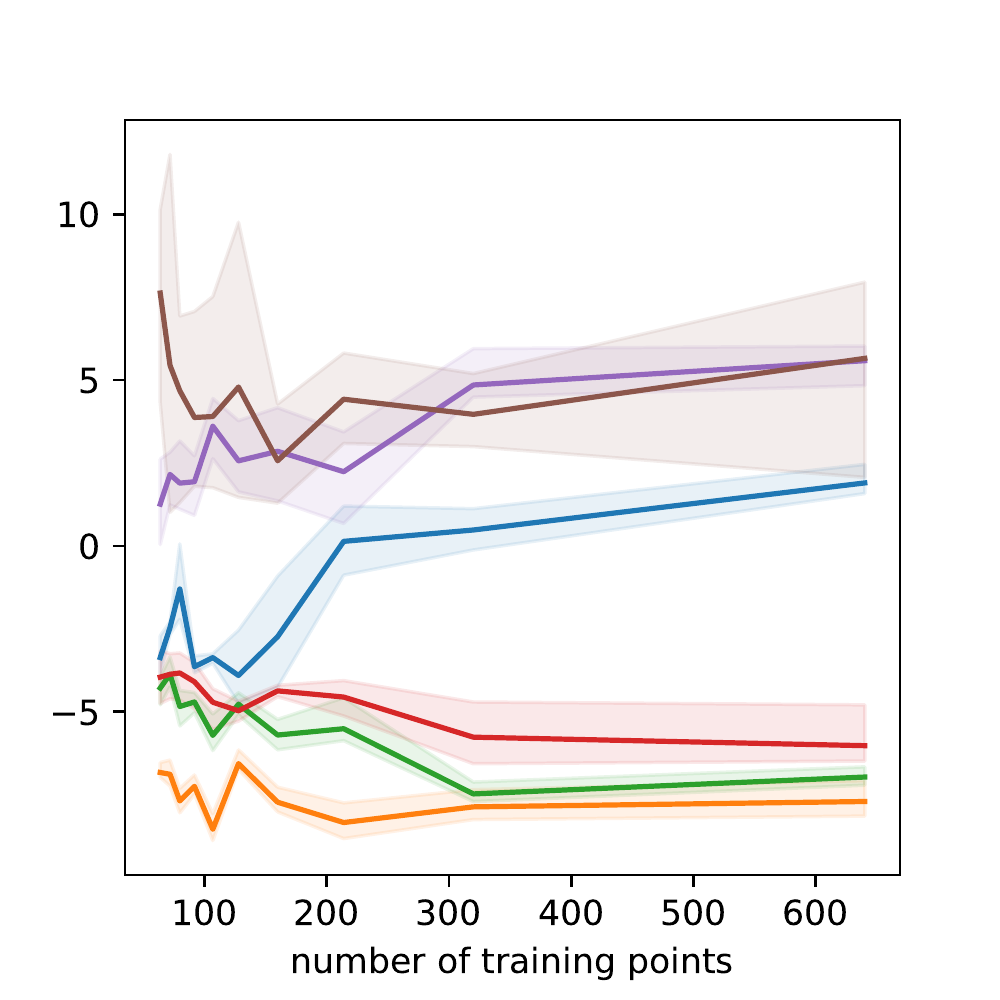}
\includegraphics[width=0.24\linewidth,trim={1mm 2mm 10mm 12mm},clip]{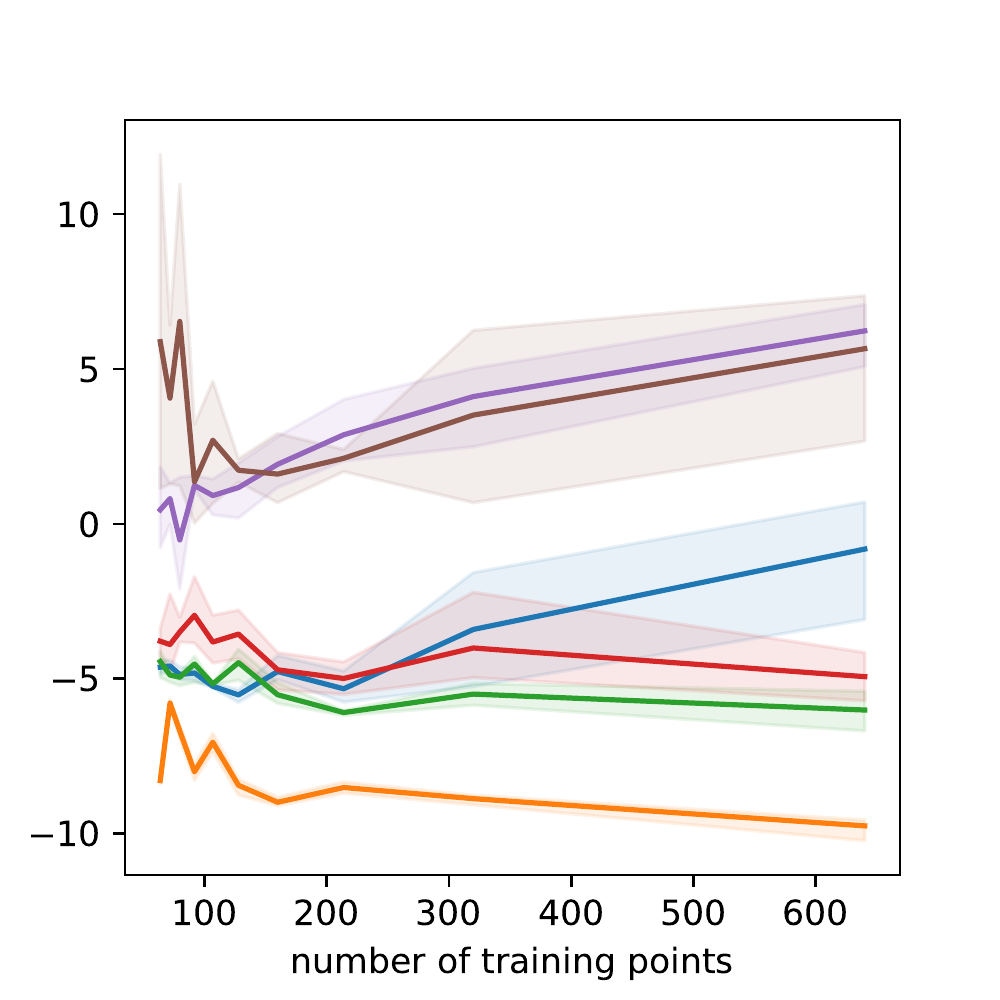}

\caption{Log-mean-squared errors between the estimated trajectories and the ground truth as a function of the sampling rate on clean (first row) and noisy (second row) observations.}
\label{fig:rate_effect}
\end{figure}

In Fig.~\ref{fig:rate_effect}, we present the effect of the sampling rate on the log mean-squared error (log-MSE) of the estimated trajectories compared to the ground truth. We show the average log-MSE for each model across five runs with a solid line, while the shaded interval represent the minimum and maximum log-MSEs among the five runs.\footnote{Minimum and maximum errors are used
to emphasize extreme cases (e.g., failed runs).}
The $x$-axis corresponds to the number of points in the training set: a smaller number of points in the training set corresponds to a lower sampling rate.
As can be seen from the results, the proposed method performs similarly to the baselines on clean observations and it outperforms the baselines on noisy observations.
We attribute the improvement to the energy conservation bias in-built in the model and to the noise filtration capabilities of the solution network.
The performance of DHH and DHPMs is more stable in the low sampling rate regime compared to the approaches with traditional numerical integrators due to computational challenges of numerical integration in this scenario. However, methods, such as Neural ODE and HNN perform better with increase of sampling rates of clean observations.
Note that the HNN model that has access to the simulator derivatives is an unrealistic approach because those derivatives are not available in practical applications.

\begin{figure}[ht]
\centering
\begin{minipage}{.24\linewidth}
\hspace{5mm}
\centering mass-spring
\end{minipage}
\hfill
\begin{minipage}{.24\linewidth}
\hspace{3mm}
\centering pendulum
\end{minipage}
\hfill
\begin{minipage}{.24\linewidth}
\hspace{3mm}
\centering 2-body system
\end{minipage}
\hfill
\begin{minipage}{.24\linewidth}
\hspace{2mm}
\centering 3-body system
\end{minipage}
\\
clean observations
\\[1mm]
\includegraphics[width=0.24\linewidth,trim={1mm 11mm 10mm 12mm},clip]{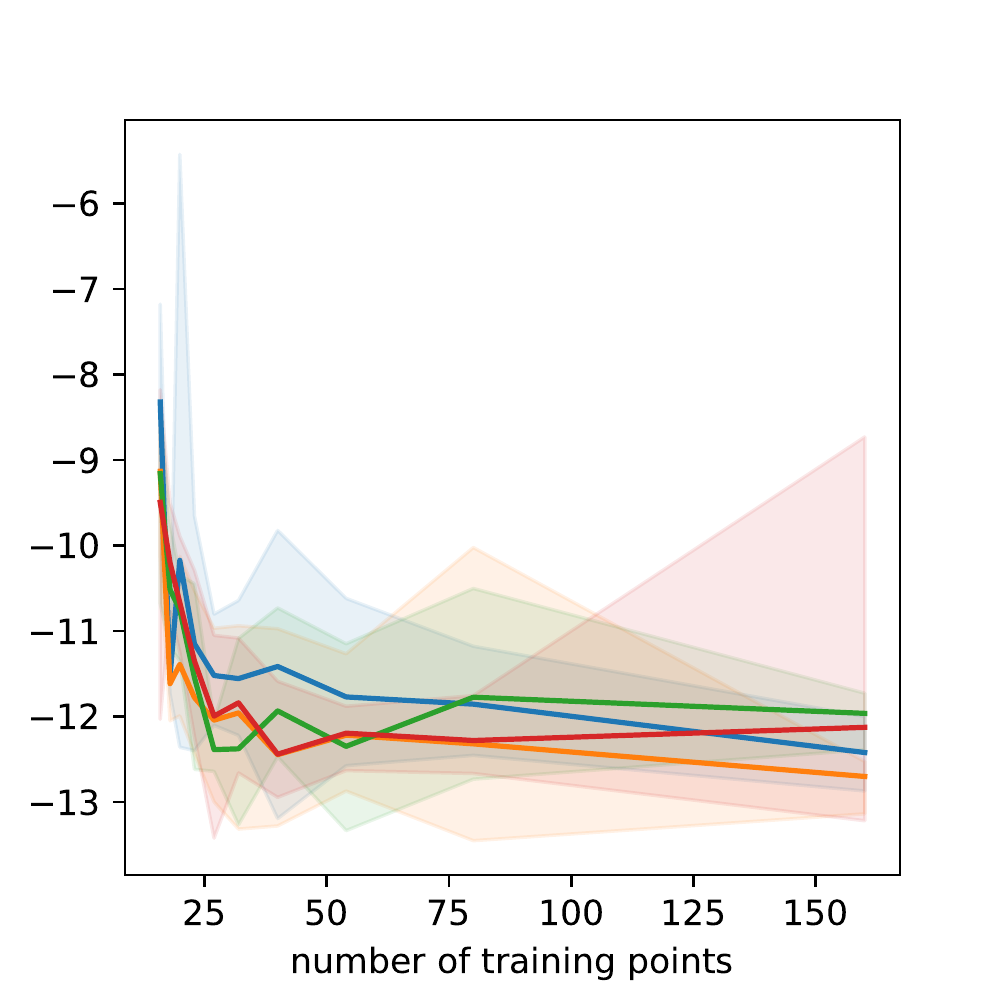}
\includegraphics[width=0.24\linewidth,trim={1mm 11mm 10mm 12mm},clip]{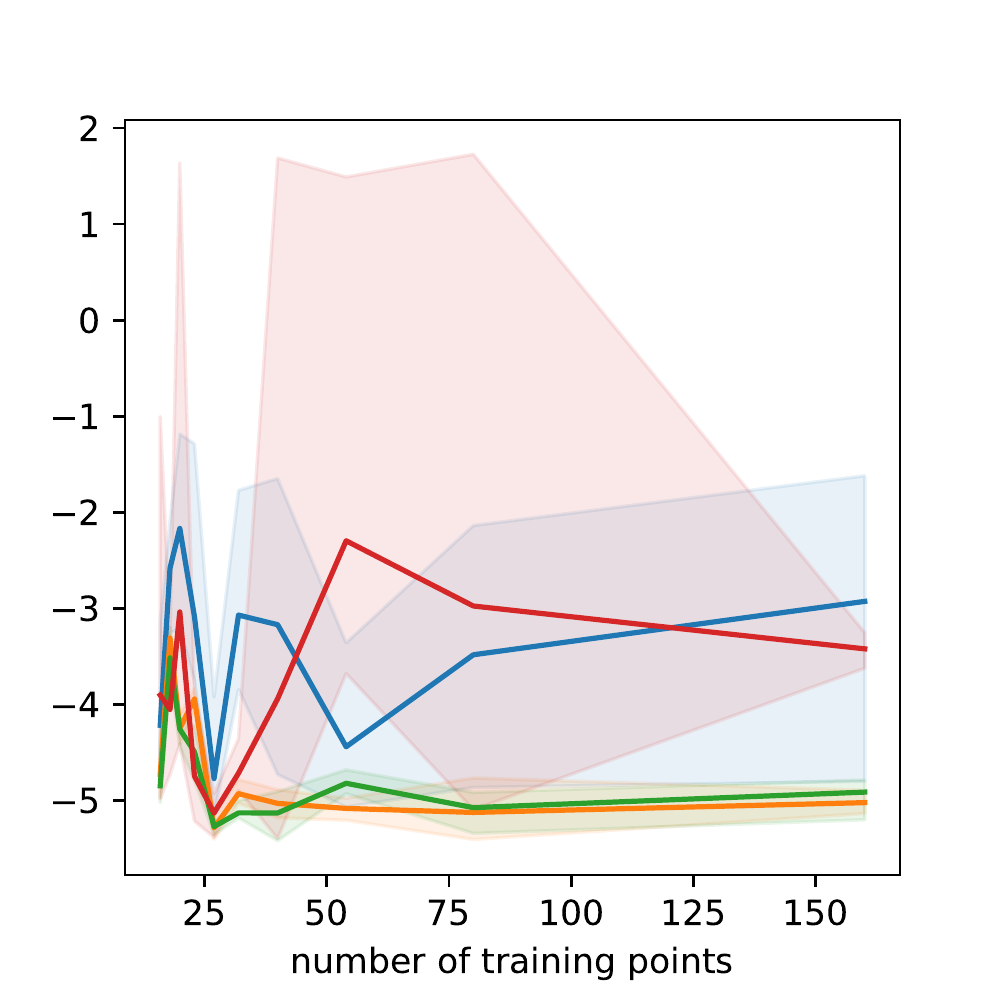}
\includegraphics[width=0.24\linewidth,trim={1mm 11mm 10mm 12mm},clip]{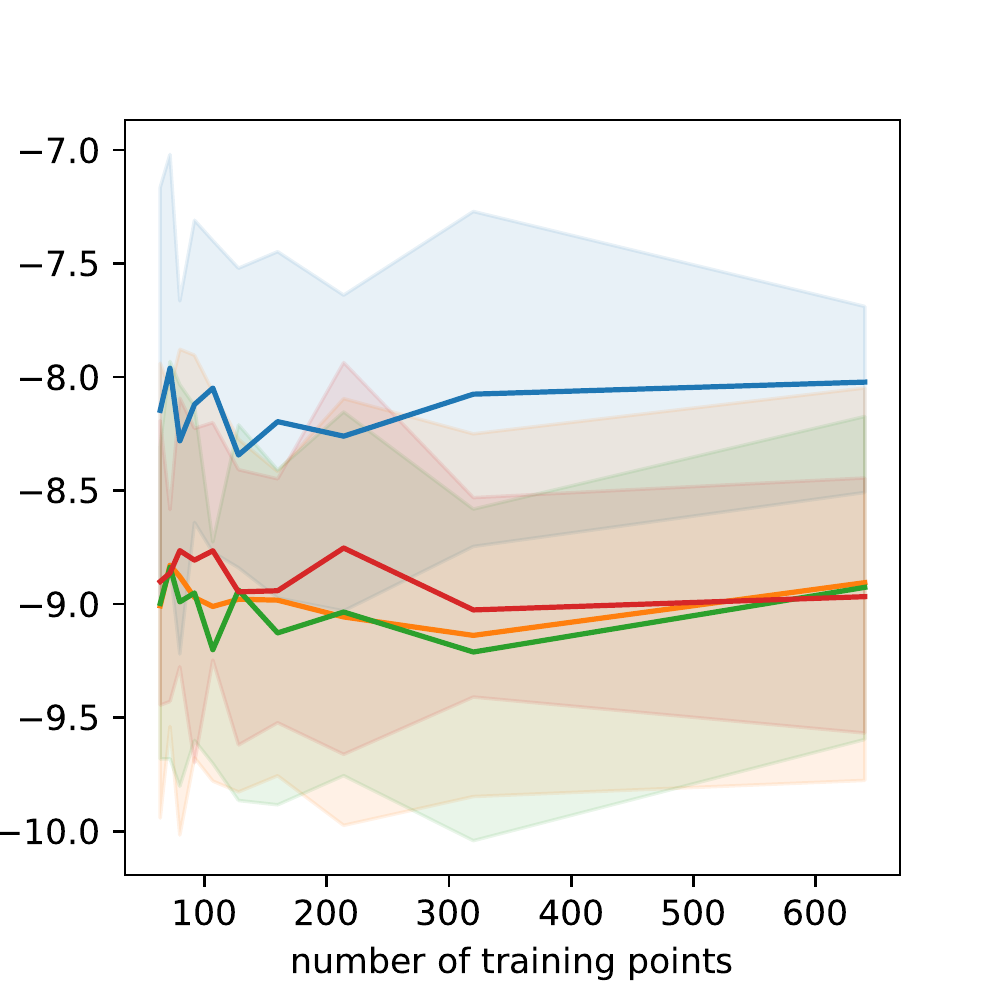}
\includegraphics[width=0.24\linewidth,trim={1mm 11mm 10mm 12mmm},clip]{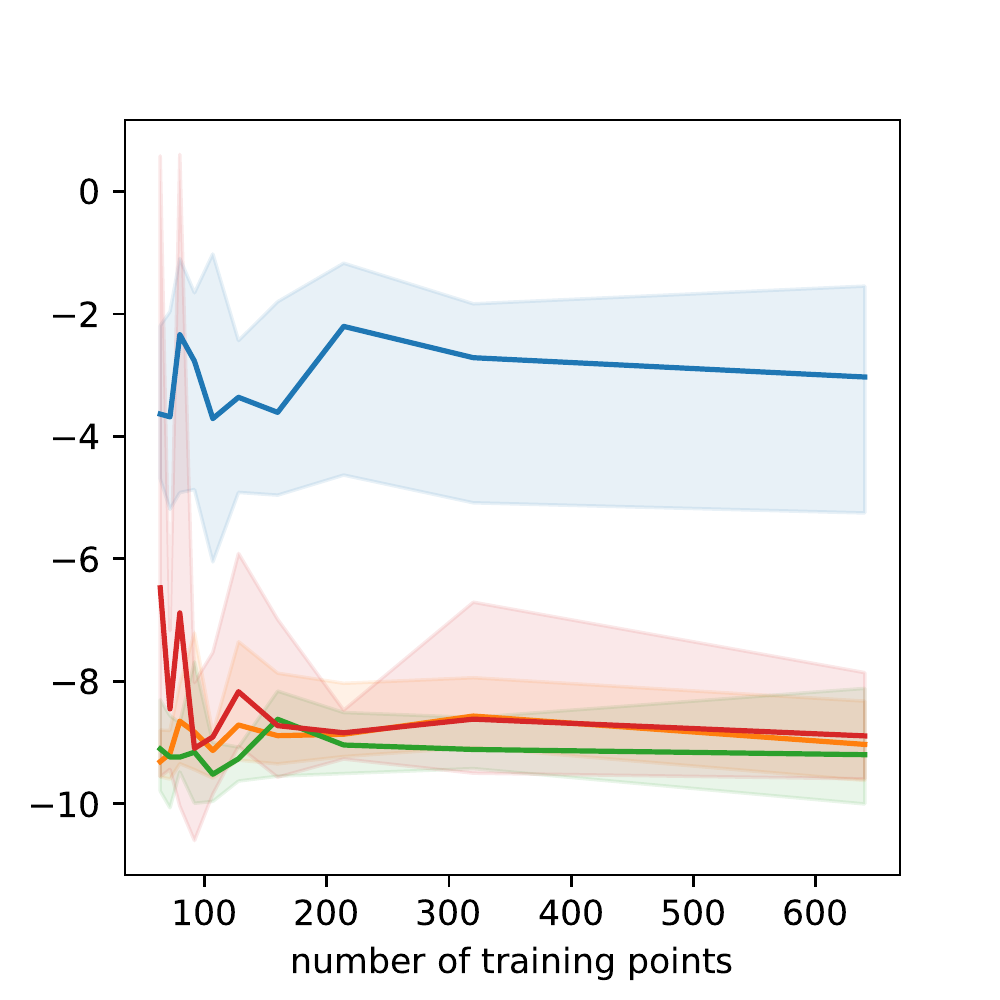}
\\
noisy observations
\\
\includegraphics[width=0.24\linewidth,trim={1mm 2mm 10mm 12mm},clip]{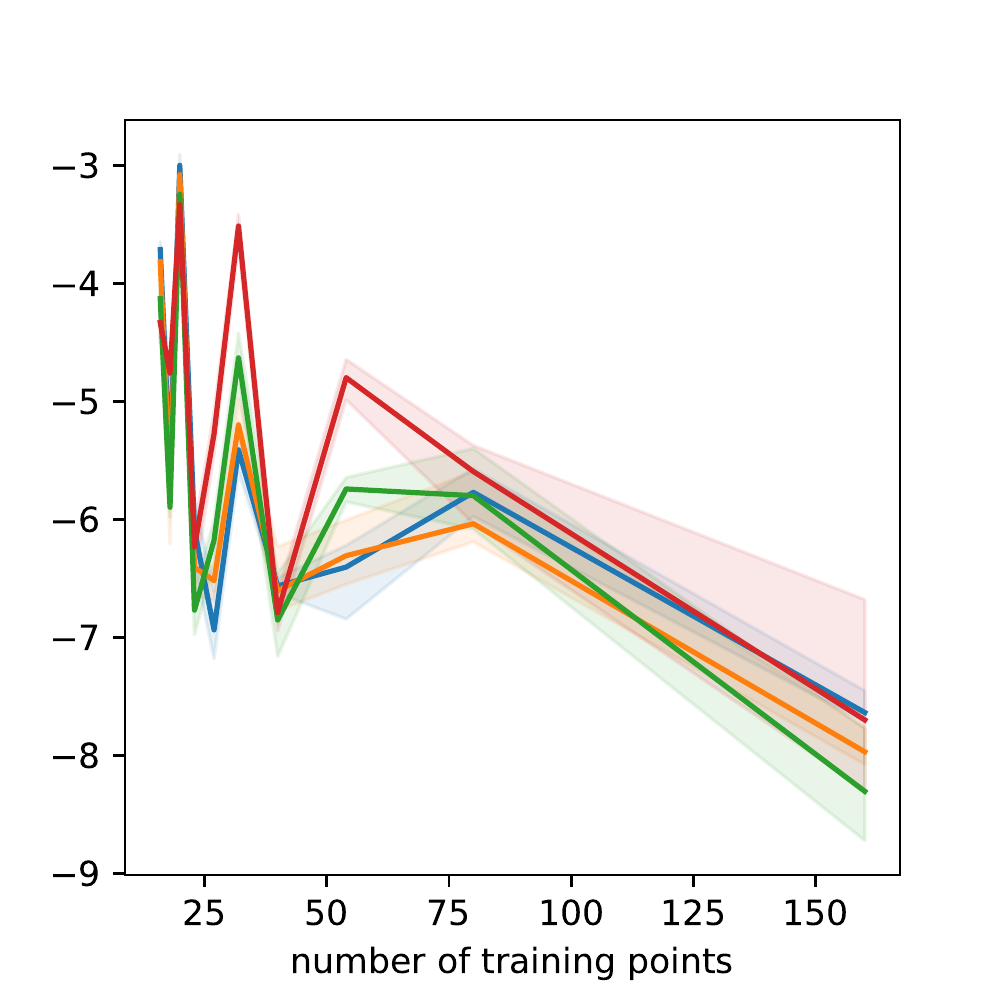}
\includegraphics[width=0.24\linewidth,trim={1mm 2mm 10mm 12mm},clip]{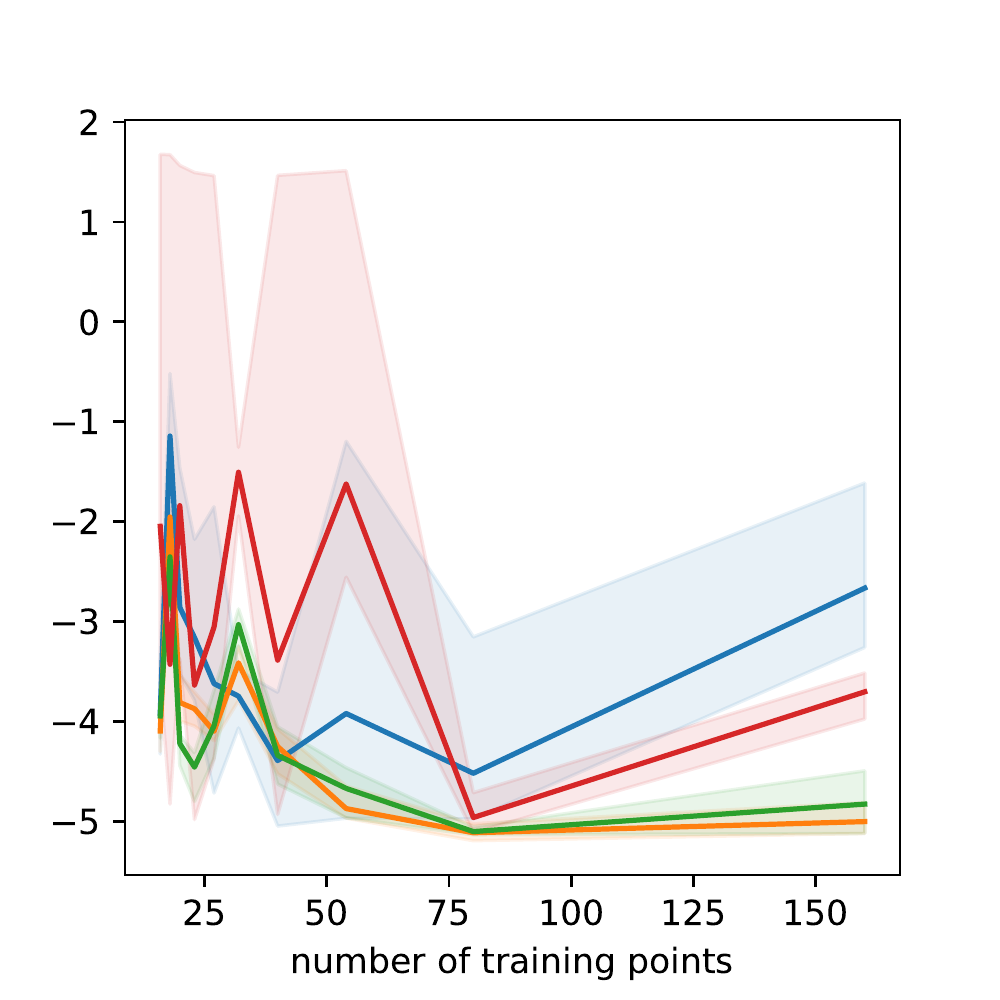}
\includegraphics[width=0.24\linewidth,trim={1mm 2mm 10mm 12mm},clip]{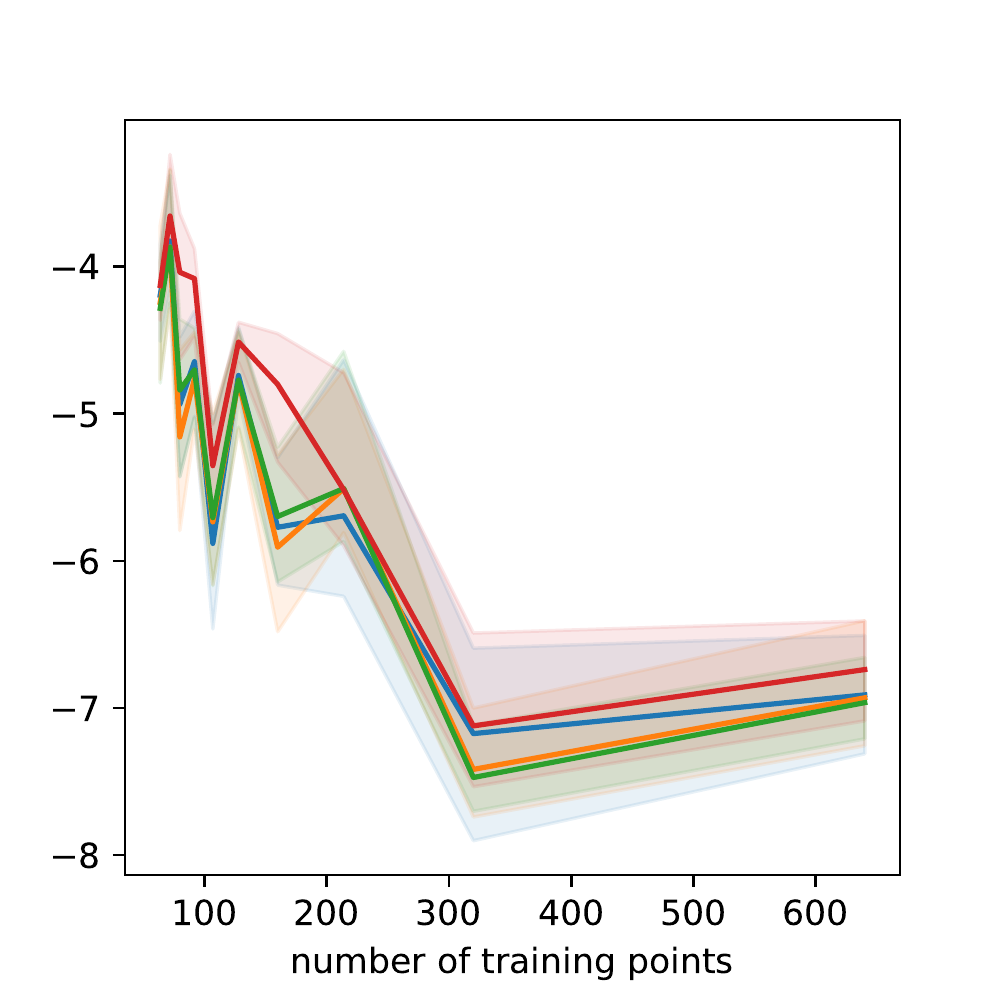}
\includegraphics[width=0.24\linewidth,trim={1mm 2mm 10mm 12mm},clip]{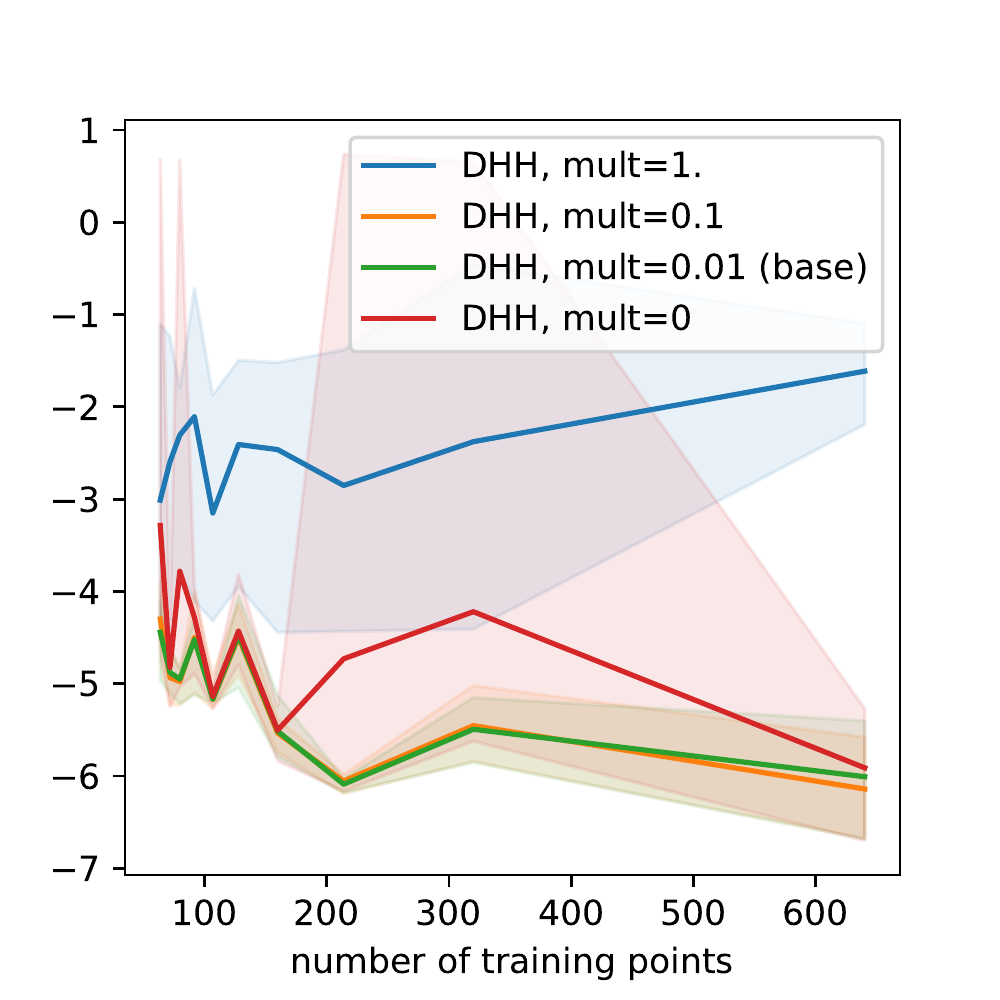}

\caption{Log-mean-squared errors between the estimated trajectories and the ground truth as a function of the sampling rate on clean (first row) and noisy (second row) observations for different multipliers for $\mathcal{L}_\text{extra}$.}
\label{fig:ablation}
\end{figure}

In Fig.~\ref{fig:ablation}, we show that the use of the extra loss term in \eqref{eq:lossextra} has a positive effect on the training of the proposed model.

\section{Conclusion and future work}

In with work, we proposed to learn a continuous-time trajectory of a modeled Hamiltonian system using an additional neural network. The time derivatives provided by this network can replace the finite-difference estimates of the derivatives used as the targets in the original HNN model. We showed experimentally that the proposed approach can outperform existing alternatives, especially in the case of low sampling rates and presence of noise in the state measurements. %

A big limitations of the HNN methodology is its applicability only to conservative systems described by Hamilton's equations and observed in the canonical coordinates. These assumptions may be restrictive in many practical applications. Addressing this limitation is an important line for future research with many promising results obtained recently (see, e.g., \cite{cranmer2020lagrangian,choudhary2021forecasting,chen2021neuralsymplectic,hoedt2021mclstm}).
Another important research question is to identify the most efficient way to incorporate inductive biases from physics into modeling of dynamical systems, which is a topic of active debate at the moment \cite{botev2021whichpriors,gruver2022deconstructing}.

\subsubsection*{Acknowledgments}
We thank CSC (IT Center for Science, Finland) for computational resources and the Academy of Finland for the support within the Flagship programme: Finnish Center for Artificial Intelligence (FCAI).

\bibliographystyle{splncs04}
\bibliography{source}

\end{document}

%% file: math_commands.tex
\usepackage{amsmath,amsfonts,bm}

\def\eqref#1{equation~\ref{#1}}

\def\1{\bm{1}}

\def\vf{{\bm{f}}}

\def\vp{{\bm{p}}}
\def\vq{{\bm{q}}}

\def\vs{{\bm{s}}}

\DeclareMathAlphabet{\mathsfit}{\encodingdefault}{\sfdefault}{m}{sl}
\SetMathAlphabet{\mathsfit}{bold}{\encodingdefault}{\sfdefault}{bx}{n}

%% file: main.bbl
\begin{thebibliography}{10}
\providecommand{\url}[1]{\texttt{#1}}
\providecommand{\urlprefix}{URL }
\providecommand{\doi}[1]{https://doi.org/#1}

\bibitem{botev2021whichpriors}
Botev, A., Jaegle, A., Wirnsberger, P., Hennes, D., Higgins, I.: Which priors
  matter? benchmarking models for learning latent dynamics. arXiv preprint
  arXiv:2111.05458  (2021)

\bibitem{chen2021gfnn}
Chen, R., Tao, M.: Data-driven prediction of general hamiltonian dynamics via
  learning exactly-symplectic maps. In: Meila, M., Zhang, T. (eds.) Proceedings
  of the 38th International Conference on Machine Learning. Proceedings of
  Machine Learning Research, vol.~139, pp. 1717--1727. PMLR (18--24 Jul 2021)

\bibitem{chen2018neuralode}
Chen, R.T.Q., Rubanova, Y., Bettencourt, J., Duvenaud, D.K.: Neural ordinary
  differential equations. In: Bengio, S., Wallach, H., Larochelle, H., Grauman,
  K., Cesa-Bianchi, N., Garnett, R. (eds.) Advances in Neural Information
  Processing Systems. vol.~31. Curran Associates, Inc. (2018)

\bibitem{chen2021neuralsymplectic}
Chen, Y., Matsubara, T., Yaguchi, T.: Neural symplectic form: Learning
  hamiltonian equations on general coordinate systems. In: Beygelzimer, A.,
  Dauphin, Y., Liang, P., Vaughan, J.W. (eds.) Advances in Neural Information
  Processing Systems (2021)

\bibitem{chen2020srnn}
Chen, Z., Zhang, J., Arjovsky, M., Bottou, L.: Symplectic recurrent neural
  networks. In: International Conference on Learning Representations (2020)

\bibitem{choudhary2021forecasting}
Choudhary, A., Lindner, J.F., Holliday, E.G., Miller, S.T., Sinha, S., Ditto,
  W.L.: Forecasting hamiltonian dynamics without canonical coordinates.
  Nonlinear Dynamics  \textbf{103}(2),  1553--1562 (2021)

\bibitem{cranmer2020lagrangian}
Cranmer, M., Greydanus, S., Hoyer, S., Battaglia, P., Spergel, D., Ho, S.:
  Lagrangian neural networks. arXiv preprint arXiv:2003.04630  (2020)

\bibitem{david2021shnn}
David, M., M{\'e}hats, F.: Symplectic learning for hamiltonian neural networks.
  arXiv preprint arXiv:2106.11753  (2021)

\bibitem{dipietro2020ssinn}
DiPietro, D., Xiong, S., Zhu, B.: Sparse symplectically integrated neural
  networks. In: Larochelle, H., Ranzato, M., Hadsell, R., Balcan, M.F., Lin, H.
  (eds.) Advances in Neural Information Processing Systems. vol.~33, pp.
  6074--6085. Curran Associates, Inc. (2020)

\bibitem{finzi2020simplifying}
Finzi, M., Wang, K.A., Wilson, A.G.: Simplifying hamiltonian and lagrangian
  neural networks via explicit constraints. In: Larochelle, H., Ranzato, M.,
  Hadsell, R., Balcan, M.F., Lin, H. (eds.) Advances in Neural Information
  Processing Systems. vol.~33, pp. 13880--13889. Curran Associates, Inc. (2020)

\bibitem{greydanus2019hamiltonian}
Greydanus, S., Dzamba, M., Yosinski, J.: Hamiltonian neural networks. In:
  Advances in Neural Information Processing Systems. vol.~32 (2019)

\bibitem{gruver2022deconstructing}
Gruver, N., Finzi, M.A., Stanton, S.D., Wilson, A.G.: Deconstructing the
  inductive biases of hamiltonian neural networks. In: International Conference
  on Learning Representations (2022)

\bibitem{hochlehnert2021physicallystruct}
Hochlehnert, A., Terenin, A., Saemundsson, S., Deisenroth, M.: Learning contact
  dynamics using physically structured neural networks. In: Banerjee, A.,
  Fukumizu, K. (eds.) Proceedings of The 24th International Conference on
  Artificial Intelligence and Statistics. Proceedings of Machine Learning
  Research, vol.~130, pp. 2152--2160. PMLR (13--15 Apr 2021)

\bibitem{hoedt2021mclstm}
Hoedt, P., Kratzert, F., Klotz, D., Halmich, C., Holzleitner, M., Nearing, G.,
  Hochreiter, S., Klambauer, G.: {MC-LSTM:} mass-conserving {LSTM}. In: Meila,
  M., Zhang, T. (eds.) Proceedings of the 38th International Conference on
  Machine Learning, {ICML} 2021. Proceedings of Machine Learning Research,
  vol.~139, pp. 4275--4286. {PMLR} (2021)

\bibitem{jagtap2020conservative}
Jagtap, A.D., Kharazmi, E., Karniadakis, G.E.: Conservative physics-informed
  neural networks on discrete domains for conservation laws: Applications to
  forward and inverse problems. Computer Methods in Applied Mechanics and
  Engineering  \textbf{365},  113028 (2020)

\bibitem{jin2020sympnets}
Jin, P., Zhang, Z., Zhu, A., Tang, Y., Karniadakis, G.E.: Sympnets: Intrinsic
  structure-preserving symplectic networks for identifying hamiltonian systems.
  Neural Networks  \textbf{132},  166--179 (2020)

\bibitem{kingma2017adam}
Kingma, D.P., Ba, J.: Adam: A method for stochastic optimization. arXiv
  preprint arXiv:1412.6980  (2014)

\bibitem{lagaris1998artificial}
Lagaris, I.E., Likas, A., Fotiadis, D.I.: Artificial neural networks for
  solving ordinary and partial differential equations. IEEE transactions on
  neural networks  \textbf{9}(5),  987--1000 (1998)

\bibitem{Lee2021deepconservation}
Lee, K., Carlberg, K.T.: Deep conservation: A latent-dynamics model for exact
  satisfaction of physical conservation laws. Proceedings of the AAAI
  Conference on Artificial Intelligence  \textbf{35}(1),  277--285 (May 2021)

\bibitem{lee2021identifying}
Lee, S., Yang, H., Seong, W.: Identifying physical law of hamiltonian systems
  via meta-learning. In: International Conference on Learning Representations
  (2021)

\bibitem{li2021hamnet}
Li, Z., Yang, S., Song, G., Cai, L.: Hamnet: Conformation-guided molecular
  representation with hamiltonian neural networks. In: International Conference
  on Learning Representations (2021)

\bibitem{mattheakis2020hamiltonian}
Mattheakis, M., Sondak, D., Dogra, A.S., Protopapas, P.: Hamiltonian neural
  networks for solving differential equations. arXiv preprint arXiv:2001.11107
  (2020)

\bibitem{raissi2018dhp}
Raissi, M.: Deep hidden physics models: Deep learning of nonlinear partial
  differential equations. The Journal of Machine Learning Research
  \textbf{19}(1),  932--955 (2018)

\bibitem{raissi2019pinn}
Raissi, M., Perdikaris, P., Karniadakis, G.E.: Physics-informed neural
  networks: A deep learning framework for solving forward and inverse problems
  involving nonlinear partial differential equations. Journal of Computational
  Physics  \textbf{378},  686--707 (2019)

\bibitem{tong2021taylornets}
Tong, Y., Xiong, S., He, X., Pan, G., Zhu, B.: Symplectic neural networks in
  taylor series form for hamiltonian systems. Journal of Computational Physics
  \textbf{437},  110325 (2021)

\bibitem{toth2020hamiltonian}
Toth, P., Rezende, D.J., Jaegle, A., Racanière, S., Botev, A., Higgins, I.:
  Hamiltonian generative networks. In: International Conference on Learning
  Representations (2020)

\bibitem{wu2020structure}
Wu, K., Qin, T., Xiu, D.: Structure-preserving method for reconstructing
  unknown hamiltonian systems from trajectory data. SIAM Journal on Scientific
  Computing  \textbf{42}(6),  A3704--A3729 (2020)

\bibitem{xiong2021nssnn}
Xiong, S., Tong, Y., He, X., Yang, S., Yang, C., Zhu, B.: Nonseparable
  symplectic neural networks. In: International Conference on Learning
  Representations (2021)

\bibitem{zhong2020symoden}
Zhong, Y.D., Dey, B., Chakraborty, A.: Symplectic ode-net: Learning hamiltonian
  dynamics with control. In: International Conference on Learning
  Representations (2020)

\bibitem{zhong2021survey}
Zhong, Y.D., Dey, B., Chakraborty, A.: Benchmarking energy-conserving neural
  networks for learning dynamics from data. In: Proceedings of the 3rd
  Conference on Learning for Dynamics and Control. Proceedings of Machine
  Learning Research, vol.~144, pp. 1218--1229. PMLR (07 -- 08 June 2021)

\bibitem{zhong2020unsupervised}
Zhong, Y.D., Leonard, N.: Unsupervised learning of lagrangian dynamics from
  images for prediction and control. In: Larochelle, H., Ranzato, M., Hadsell,
  R., Balcan, M.F., Lin, H. (eds.) Advances in Neural Information Processing
  Systems. vol.~33, pp. 10741--10752. Curran Associates, Inc. (2020)

\bibitem{zhu2020deephnn}
Zhu, A., Jin, P., Tang, Y.: Deep hamiltonian networks based on symplectic
  integrators. arXiv preprint arXiv:2004.13830  (2020)

\end{thebibliography}
